\newif\ifRAL
\RALfalse 	

\newif\ifTR
\TRfalse 	

\newif\ifPrePrint
\PrePrinttrue

\newif\ifDraft
\Draftfalse

\ifRAL
\documentclass[letterpaper, 10 pt, journal, twoside]{IEEEtran}
\else
\documentclass[letterpaper, 10 pt, conference]{ieeeconf}
\fi

\IEEEoverridecommandlockouts

\ifRAL
\else
\fi

\makeatletter

\let\proof\@undefined
\let\endproof\@undefined
\makeatother

\usepackage{amsmath} 
\usepackage{amssymb}  
\usepackage{amsthm}
\usepackage{amsfonts}
\usepackage{mathtools}

\usepackage{bm}
\providecommand{\bm}{\pmb}

\usepackage{siunitx}
\usepackage{soul}

\theoremstyle{definition}

\theoremstyle{remark}

\usepackage{verbatim}

\usepackage{color}
\usepackage{subcaption}
\usepackage{graphicx}
\usepackage[font=small]{caption}

\usepackage{optidef}

\usepackage{times}

\usepackage{xspace}

\usepackage[ruled,vlined,linesnumbered]{algorithm2e}
\usepackage[noend]{algorithmic}

\graphicspath{{images/}}

\usepackage[us]{datetime}
\usepackage{caption}
\usepackage[usenames,dvipsnames,table]{xcolor}

\usepackage{hyperref} 
\hypersetup{
    colorlinks,
    citecolor=black,
    filecolor=black,
    linkcolor=black,
    urlcolor=black,
    pdfauthor={},
    pdfsubject={},
    pdftitle={}
}

\usepackage{todonotes}

\usepackage{graphicx}

\usepackage{latexsym}
\usepackage{color}

\usepackage{cite}

\usepackage{caption}

\usepackage{tikz}
\usetikzlibrary{shapes,arrows,calc,fit}
\tikzstyle{node} = [rectangle, text centered, draw=black, rounded corners, align=center]

\usepackage{tabularx, booktabs}
\newcolumntype{Y}{>{\centering\arraybackslash}X}
\usepackage{multirow}
\usepackage{paralist}
\usepackage{acro}       
\acsetup{single = true}
\DeclareAcronym{UAV}{
  short=UAV,
  long = unmanned aerial vehicle,
  short-indefinite = a,
  long-indefinite = an
}

\DeclareAcronym{MAV}{
  short=MAV,
  long = micro aerial vehicle,
}

\DeclareAcronym{OMAV}{
  short=OMAV,
  long = omnidirectional micro aerial vehicle,
  short-indefinite = an,
  long-indefinite = an
}

\DeclareAcronym{IMU}{
  short=IMU,
  long = inertial measurement unit,
  short-indefinite = an,
  long-indefinite = an
}

\DeclareAcronym{RC}{
  short=RC,
  long = remote control,
  short-indefinite = an,
  long-indefinite = a
}

\DeclareAcronym{FPV}{
  short=FPV,
  long = first person view,
  short-indefinite = an,
  long-indefinite = a
}

\DeclareAcronym{MOCAP}{
  short=MOCAP,
  long = motion capture system,
  short-indefinite = a,
  long-indefinite = a
}

\usepackage{color}

\newcommand{\add}[1]{#1}
\newcommand{\remove}[1]{}

\newcommand{\vect}[1]{\bm{#1}}		
\newcommand{\matr}[1]{\bm{#1}}		
\newcommand{\nR}[1]{\mathbb{R}^{#1}}		
\newcommand{\SO}[1]{\mathsf{SO}(#1)}		
\newcommand{\SE}[1]{\mathsf{SE}(#1)}		
\newcommand{\matrice}[1]{\begin{bmatrix} #1 \end{bmatrix}}	
\newcommand{\upperRomannumeral}[1]{\uppercase\expandafter{\romannumeral#1}}	

\newcommand{\vecInFrame}[2]{#1^{#2}}

\newcommand{\transp}{^\top}

\newcommand{\fig}{Fig.~}	

\renewcommand{\frame}{\mathcal{F}}		
\newcommand{\origin}{O}						
\newcommand{\vX}{\vect{x}}					
\newcommand{\vY}{\vect{y}}					
\newcommand{\vZ}{\vect{z}}					
\newcommand{\pos}{\vect{p}}				
\newcommand{\dpos}{\vect{v}}				
\newcommand{\rotMat}{\matr{R}}				
\newcommand{\vZero}{\vect{0}}				
\newcommand{\eye}[1]{\matr{I}_{#1}}

\newcommand{\frameW}{\frame_W}			
\newcommand{\frameWAbb}{W}			
\newcommand{\originW}{\origin_W}		
\newcommand{\xW}{\vX_W}				
\newcommand{\yW}{\vY_W}				
\newcommand{\zW}{\vZ_W}				

\newcommand{\frameM}{\frame_M}		
\newcommand{\originM}{\origin_M}	
\newcommand{\xM}{\vX_M}				
\newcommand{\yM}{\vY_M}				
\newcommand{\zM}{\vZ_M}				

\newcommand{\frameH}{\frame_H}		
\newcommand{\frameHAbb}{H}		
\newcommand{\originH}{\origin_H}	
\newcommand{\xH}{\vX_H}				
\newcommand{\yH}{\vY_H}				
\newcommand{\zH}{\vZ_H}				

\newcommand{\frameS}{\frame_S}		
\newcommand{\frameSAbb}{S}			
\newcommand{\frameRefS}{\frame_{S,ref}}		
\newcommand{\frameRefSAbb}{{S,ref}}		
\newcommand{\originS}{\origin_S}	
\newcommand{\xS}{\vX_S}				
\newcommand{\yS}{\vY_S}				
\newcommand{\zS}{\vZ_S}				

\newcommand{\humanPos}{\pos_H} 
\newcommand{\humanVel}{\dpos_H} 
\newcommand{\dhumanVel}{{\dot{\dpos}_H}} 
\newcommand{\humanAttQuat}{\vect{q}^M_H} 
\newcommand{\humanAtt}{{\matr{R}^M_H}} 
\newcommand{\humanRate}{\vect{\omega}_{\frameHAbb}} 
\newcommand{\dhumanRate}{{\dot{\vect{\omega}}_{\frameHAbb}}} 

\newcommand{\humanInert}{\matr{M}_H} 
\newcommand{\humanDamping}{\matr{D}_H} 
\newcommand{\humanWrenchActive}{\vect{\tau}_{H,act}} 

\newcommand{\humanAdmInertia}{{\matr{M}_{adm}}} 
\newcommand{\humanAdmInertiaTransl}{{\matr{M}_{adm,t}}} 
\newcommand{\humanAdmInertiaRot}{{\matr{M}_{adm,r}}} 
\newcommand{\humanAdmDamp}{{\matr{D}_{adm}}} 
\newcommand{\humanAdmDampTransl}{{\matr{D}_{adm,t}}} 
\newcommand{\humanAdmDampRot}{{\matr{D}_{adm,r}}} 

\newcommand{\humanWrench}{\vect{\tau}_H} 

\newcommand{\slavePos}{\pos_S} 
\newcommand{\slavePosY}{\pos_{S,y}} 
\newcommand{\slaveVel}{\dpos_S} 
\newcommand{\slaveAcc}{\dot{\dpos}_S} 
\newcommand{\slaveAtt}{{\rotMat^\frameWAbb_\frameSAbb}} 
\newcommand{\slaveRate}{\vect{\omega}_{\frameSAbb}} 
\newcommand{\slaveAngAcc}{\dot{{\vect{\omega}}}_{\frameSAbb}} 
\newcommand{\slaveExtWrench}{\hat{\vect{\tau}}_{ext}} 

\newcommand{\slaveRefPos}{\pos_{S,ref}} 
\newcommand{\slaveRefVel}{\dpos_{S,ref}} 
\newcommand{\slaveRefAtt}{{\rotMat^\frameWAbb_\frameRefSAbb}} 
\newcommand{\slaveRefRate}{\vect{\omega}_{\frameRefSAbb}} 

\newcommand{\slaveErrorPos}{\vect{e}_{p}} 
\newcommand{\slaveErrorVel}{\vect{e}_{v}} 
\newcommand{\slaveErrorAtt}{\vect{e}_{R}} 
\newcommand{\slaveErrorRate}{\vect{e}_{\omega}} 

\newcommand{\slaveVirtInert}{\matr{M}_v} 
\newcommand{\slaveVirtStiff}{\matr{K}_v} 
\newcommand{\slaveVirtDamp}{\matr{D}_v} 

\newcommand{\refgenMaxVel}{v_{max}}
\newcommand{\refgenMaxRate}{\omega_{max}}

\newcommand{\feedbackTotalWrench}{\vect{\tau}_{fb,total}} 

\newcommand{\feedbackRecenterWrench}{\vect{\tau}_{fb,rec}} 
\newcommand{\feedbackRecenterStiff}{\matr{K}_{rec}} 
\newcommand{\feedbackRecenterStiffTransl}{\matr{K}_{rec,t}} 
\newcommand{\feedbackRecenterStiffRot}{\matr{K}_{rec,r}}

\newcommand{\feedbackExtWrench}{\vect{\tau}_{fb,ext}} 
\newcommand{\contactForces}{\vect{\tau}_{c,1:3}} 
\newcommand{\contactTorques}{\vect{\tau}_{c,4:6}} 

\ifRAL 
\fi

\author{Mike Allenspach$^1$, Nicholas Lawrance$^{1}$, Marco Tognon$^{1}$, and Roland Siegwart$^1$
	\ifRAL
	\thanks{Manuscript received: DD,\,MM,\,YY; Revised DD,\,MM,\,YY ; Accepted DD,\,MM,\,YY.}
	\thanks{This paper was recommended for publication by Editor NAME SURNAME upon evaluation of the Associate Editor and Reviewers' comments.
	This work was partially funded by ...} 
	\fi
		\thanks{$^1$ Authors are with the Autonomous Systems Lab, ETH
Z\"urich, Z\"urich 8092, Switzerland {\tt \footnotesize
			\href{mailto:mike.allenspach@mavt.ethz.ch}{mike.allenspach@mavt.ethz.ch},
			\href{mailto:nicholas.lawrance@mavt.ethz.ch}{nicholas.lawrance@mavt.ethz.ch},
			\href{mailto:mtognon@ethz.ch}{mtognon@ethz.ch},
			\href{mailto:rsiegwart@ethz.ch}{rsiegwart@ethz.ch}}
	}
	\thanks{This research was in part supported by the National Center of Competence in Research (NCCR) on Digital Fabrication, in part by the NCCR Robotics, and in part by Armasuisse Science and Technology.}
	\ifRAL
	\thanks{Digital Object Identifier (DOI): see top of this page.}
	\fi
}

\ifRAL 
\title{Towards 6DoF Bilateral Teleoperation of an Omnidirectional Aerial Vehicle for Aerial Physical Interaction}

\else 
\title{\bf Towards 6DoF Bilateral Teleoperation of an Omnidirectional Aerial Vehicle for Aerial Physical Interaction}
\fi

\ifPrePrint
\makeatletter
\def\ps@titlepagestyle{
	\def\@oddfoot{}\def\@evenfoot{}
	\def\@oddhead{\textcolor{red}{\sf\footnotesize Preprint version, final version at http://ieeexplore.ieee.org/ \hfill IEEE International Conference on Robotics and Automation (ICRA) 2022}}
	\def\@evenhead{\textcolor{red}{\sf\footnotesize  Preprint version, final version at http://ieeexplore.ieee.org/  \hfill IEEE International Conference on Robotics and Automation (ICRA) 2022}}%
}%
\makeatother
\makeatletter
\def\ps@headings{
	\def\@oddfoot{\textcolor{red}{\sf\footnotesize  Preprint version, final version at http://ieeexplore.ieee.org/ \hfill \thepage \;\;~\hfill~\hfill IEEE International Conference on Robotics and Automation (ICRA) 2022}}\def\@evenfoot{\hfill\thepage\hfill}
	\def\@oddhead{}\def\@evenhead{}%
}%
\makeatother
\fi

\ifDraft
\makeatletter
\def\ps@titlepagestyle{
	\def\@oddfoot{}\def\@evenfoot{}
	\def\@oddhead{\textcolor{red}{\sf Draft version  \hfill Confidential}}
	\def\@evenhead{\textcolor{red}{\sf  Draft version  \hfill Confidential}}%
}%
\makeatother
\makeatletter
\def\ps@headings{
	\def\@oddfoot{\textcolor{red}{\sf  Draft version  \hfill Confidential}}\def\@evenfoot{\hfill\thepage\hfill}
	\def\@oddhead{}\def\@evenhead{}%
}%
\makeatother
\usepackage{draftwatermark}
\SetWatermarkText{Confidential draft}
\SetWatermarkScale{0.6}
\SetWatermarkLightness{0.9}

\fi

\begin{document}

\maketitle

\begin{abstract}
Bilateral teleoperation offers an intriguing solution towards shared autonomy with aerial vehicles in contact-based inspection and manipulation tasks.
Omnidirectional aerial robots allow for full pose operations, making them particularly attractive in such tasks.
Naturally, the question arises whether standard bilateral teleoperation methodologies are suitable for use with these vehicles.
In this work, a fully decoupled 6DoF bilateral teleoperation framework for aerial physical interaction is designed and tested for the first time. 
The method is based on the well established rate control, recentering and interaction force feedback policy. 
\remove{Specific hardware requirements to implement a 6DoF haptic feedback are highlighted.}
However, practical experiments evince the difficulty of performing decoupled motions in a single axis only.
\add{As such,} this work shows that the trivial extension of standard methods is insufficient for omnidirectional teleoperation, due to the operator's physical inability to properly decouple all input DoFs.
This suggests that further studies on enhanced haptic feedback are necessary.
\end{abstract}

\ifRAL 
\else 
{} 
\fi


%
\section{INTRODUCTION}\label{sec:intro}

\ifRAL
\IEEEPARstart{T}{he}
\else
The
\fi
physical interaction between flying robots and the environment has gained increasing interest in the robotics community in recent years.
Aerial robots with manipulation capabilities have already been successfully deployed in a variety of interaction tasks~\cite{2019e-TogTelGasSabBicMalLanSanRevCorFra,2019-BodBruPanWalPfaAngSieNie,2021-Ollero}.
The low-cost, high maneuverability and nearly unlimited workspace of these aerial manipulators allow deployment in hard-to-reach or remote places, as well as when contact-based inspection and maintenance are too dangerous for human operators~\cite{2018-Ruggiero,2020-Meng}.
In this regard, \acp{OMAV} offer a particularly compelling solution.
Their capability to generate thrust in any direction allows \add{hovering at arbitrary orientations, as well as independently controlling position and attitude}. \add{Thus, these platforms are capable of precise motion and interaction force control, while simultaneously rejecting disturbances} \cite{2021-BodTogSie,2018-ParLeeAhnKimHerYanLee, 2018-BreDAn}.
\\
Despite recent advances in autonomous control of these vehicles (e.g. \cite{2019-Ryll,2019-Trujillo,2020-Bodie}), existing regulations and safety requirements often still require a human operator in the loop.
Real-time inclusion of the human expert's knowledge is especially important when complex tasks must be performed in uncertain or a-priori unknown environments, considering the limited decisional autonomy of modern robots. 
Transferring the operator supervision and decision making skills to the remote site requires careful design of teleoperation systems.
On the one hand, taking full control of the robot and exploiting its capabilities is only possible if every DoF is individually controllable by the operator.
In view of the recent trend towards omnidirectional aerial manipulators, teleoperation frameworks must therefore simultaneously provide fully decoupled commands for all three translational and rotational axes.
On the other hand, meaningful system information must be reflected back to the operator through haptic and/or visual feedback.
This concept of \textit{bilateral teleoperation} improves the user situational awareness, which in turn supports their decision making process~\cite{2014-MerStrCar}.

\remove{Different types of motion generation and feedback policies for bilateral teleoperation of aerial robots have already been well established in the community.}
\remove{However, s}State-of-the-art \add{bilateral teleoperation} approaches are almost exclusively focused on underactuated platforms for which the operator can control the position and yaw angle only.
Naturally, the question arises if an extension to omnidirectional vehicles is straightforward or if additional considerations and/or problems must be addressed.
Specifically, the goal of this work is to evaluate whether it is possible to teleoperate an \ac{OMAV} using standard methodologies, in both contact-less and contact-based conditions.
%
%

\subsection{Related Work}
Early works on teleoperation methods for aerial robots primarily considered contact-free flight rather than interaction~\cite{2010-Schill,2011-Rifai}, mostly focusing on direct control of the vehicles and the obstacle avoidance problem.
An alternative teleoperation strategy was suggested in~\cite{2018-Masone}, where the operator indirectly steers the \ac{MAV} by modifying the parameters of a dynamic path.
Hereby, user feedback includes information about tracking performance or the presence of obstacles.


%

Only recently, bilateral teleoperation of \acp{MAV} has been extended to aerial physical interaction as well.
One of the earliest work to apply bilateral teleoperation to aerial physical interaction has been presented in \cite{2015-Gioioso} but was restricted to simulation.
The authors proposed the use of a haptic device with three actuated translational DoFs to command both the motion of the vehicle and the interaction force when in contact, mapping the position of the input device into the desired acceleration of the \ac{MAV}.
Simultaneously, the device renders a feedback force aimed to recenter the input device, as well as to provide an indication of the measured interaction force.
A similar reference and feedback generation scheme is used in~\cite{2019-Islam}, although environmental forces are estimated using a risk field interaction model.
Relying on standard rate control and interaction force feedback, the framework in \cite{2020-Lee,2020-Coelho} makes use of passivity theory to ensure stability even under communication delays.

The methods presented thus far focus exclusively on underactuated platforms and are naturally limited to the position control of the vehicle.
In fact, \cite{2018-ParLeeAhnKimHerYanLee} is the only work where the topic of omnidirectional bilateral teleoperation for aerial robots is addressed.
However, since it is not the main contribution of the article, specifics regarding the implementation of reference and feedback generation schemes, as well as detailed discussions are missing.
Furthermore, although the framework seems to support decoupled 6DoF bilateral teleoperation in theory, the experimental verification is limited to translational motion only.

Even though the extension of well established methods may appear trivial, a detailed evaluation is still missing.
It is important to note that standard solutions developed for ground manipulators (e.g. \textit{Leader-Follower-Configuration}) are unsuitable for aerial robotics, since the limited input workspace must be mapped to a virtually unlimited robot workspace.
Additional issues arise due to the complexity of the $\SE{3}$ control space, containing an open vector space for translation and a closed isometry for orientation.
In summary, effective bilateral teleoperation for \acp{OMAV} remains an unsolved problem, let alone its application in physical aerial interaction tasks.




\subsection{Contributions}
\add{In summary,} the main contributions of this work are\remove{ summarized as follows}:
\begin{itemize}
    \item Design of a fully decoupled 6DoF teleoperation framework by \add{extending} the \remove{well} established rate control, recentering and interaction force feedback policy to \add{$\SE{3}$}.
    \item Evaluation of the proposed policy in real-world flight experiments, including free-flight omnidirectional reference generation, as well as push-and-slide operation during physical interaction.
    \item Discussion about limitations of standard policies, namely the operator's physical inability to properly decouple all input DoFs.
\end{itemize}
As such, this study serves as a first step towards remote controlled omnidirectional aerial physical interaction, by identifying working features and potential issues when adapting standard methods used for underactuated \acp{MAV}.
\begin{figure}[t]
    \centering
    \begin{subfigure}[b]{.9\linewidth}
        \includegraphics[trim={0 4cm 0 1cm},clip,width=\linewidth]{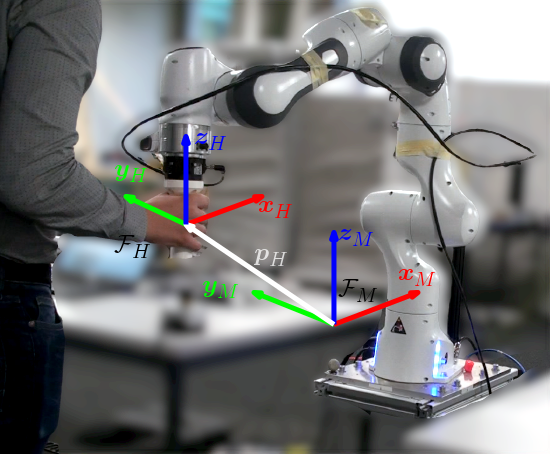}%
        \caption{Local environment: human and haptic device.}
        \label{fig:frames_local}
    \end{subfigure}\vspace{0.25cm}
    \begin{subfigure}[b]{.9\linewidth}
        \includegraphics[trim={0 2cm 0 2cm},clip,width=\linewidth]{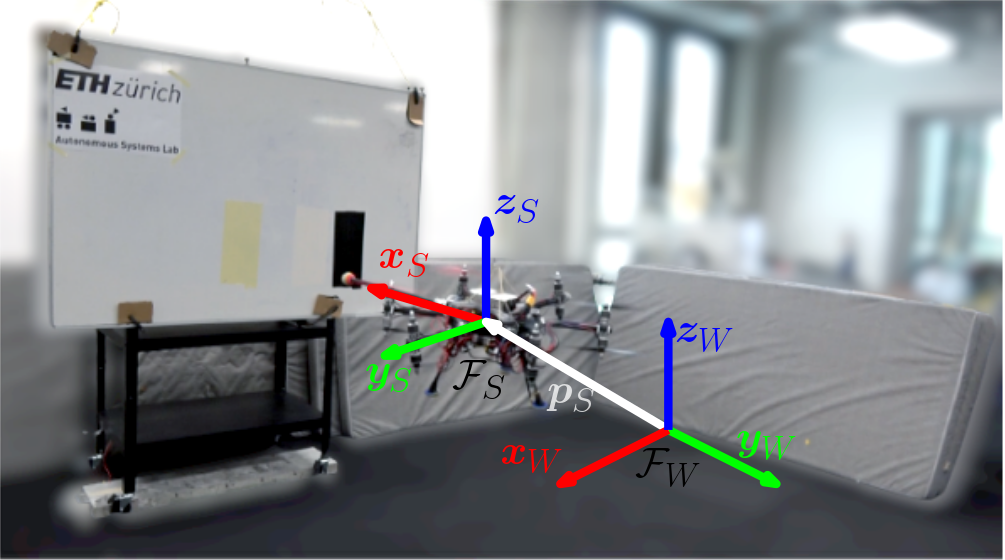}
        \caption{Remote work environment: aerial robot and task objects.}
        \label{fig:frames_remote}
    \end{subfigure}
    \caption{Representation the 6DoF bilateral teleoperation setup.}
    \label{fig:frames}
\end{figure}

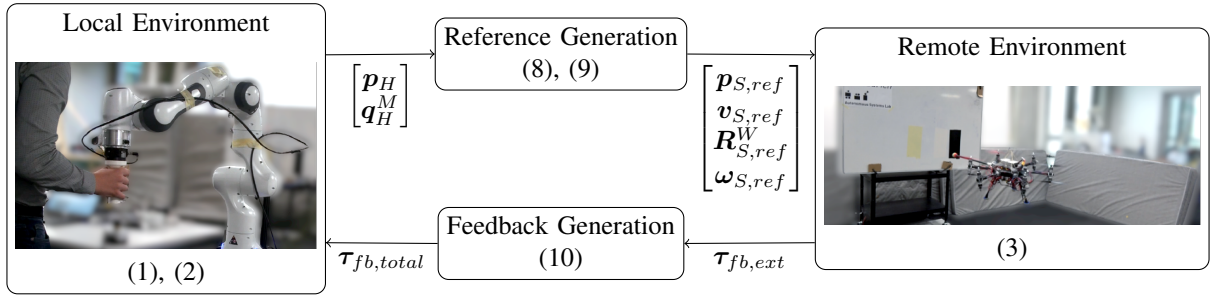
\begin{figure*}[htb!]
    \begin{center}
        \begin{tikzpicture}
            \node[node] (panda) {Local Environment\\\\\includegraphics[trim={0 4cm, 0 0cm},clip,width=4.0cm]{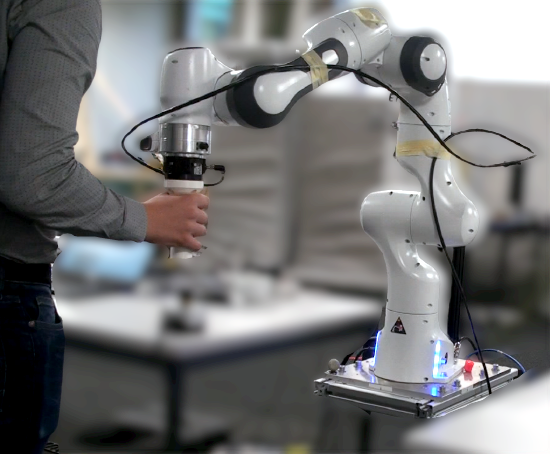}\\\eqref{eq:human_model},\ \eqref{eq:panda_model}};
            \node[right=3cm of panda](aux){};
            \node[node, above = 0.65cm of aux] (refgen) {Reference Generation\\\eqref{eq:ref_gen_transl}, \eqref{eq:ref_gen_rot}};

            \node[node, below = 0.65cm of aux] (fbgen) {Feedback Generation\\\eqref{eq:fb_gen_tot}};

            \node[node, right = 3.25cm of aux] (omav) {Remote Environment\\\\\includegraphics[trim={0 2cm 0 2cm},clip,width=5.0cm]{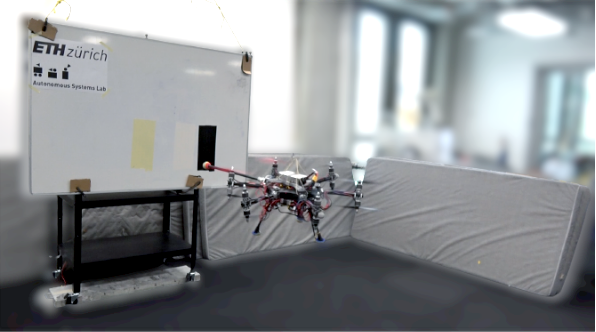}\\\eqref{eq:omav_impedance}};


            \draw[->] ($(panda.east)+(0,1.25)$)--node[below]{$\matrice{\humanPos\\\humanAttQuat}$} (refgen.west);
            \draw[->] (refgen.east)--node[below]{$\matrice{\slaveRefPos\\\slaveRefVel\\\slaveRefAtt\\\slaveRefRate}$} ($(omav.west)+(0,1.25)$);

            \draw[->] (fbgen.west) -- node[below]{$\feedbackTotalWrench$}($(panda.east)+(0,-1.25)$);

            \draw[->] ($(omav.west)+(0,-1.25)$)--node[below]{$\feedbackExtWrench$} (fbgen.east);

%
        \end{tikzpicture}
    \end{center}
    \caption{Detailed interactions between the different components of the proposed teleoperation framework.}
    \label{fig:system_design_detail}
\end{figure*}
\section{Modeling}\label{sec:model}
The system considered in a bilateral teleoperation framework consists of a \textit{human operator}, a \textit{haptic device}, and the \textit{robot} (in our case \iac{OMAV}).
Human and haptic device constitute the local environment and are connected through a virtual communication link to the robot in the remote work environment (see Fig.~\ref{fig:frames}).
The human operator is in constant contact with the handle of the haptic device which allows them to perform small-scale translational and rotational perturbations.
Hereby, the mechanical construction and actuation of the input device must support decoupled 6DoF motion and \remove{wrench}\add{force/torque} feedback rendering.

To describe the local configuration, we define the inertial frame $\frameM=\{\originM,\xM,\yM,\zM\}$ with origin $\originM$ and unit axes $\{\xM,\yM,\zM\}$ corresponding to the idle pose of the haptic device's handle.
Its current position and orientation are captured by the frame $\frameH=\{\originH,\xH,\yH,\zH\}$ with origin $\originH$ rigidly attached to the handle.
In particular, it will \add{become} clear that when $\frameH$ and $\frameM$ coincide, the desired rate commanded to the aerial robot is zero.
In a similar fashion, two additional frames are defined for the remote environment.
The inertial world frame, $\frameW=\{\originW,\xW,\yW,\zW\}$ is located at an arbitrary origin point $\originW$, such that $\zW$ is opposite to gravity.
Finally, the state of the aerial vehicle is described by the body frame $\frameS=\{\originS,\xS,\yS,\zS\}$ whose origin $\originS$ coincides with the \ac{OMAV}’s center-of-mass and $\xS$ points along the end-effector used during physical contact.
A right-hand superscript, e.g. $\vecInFrame{\vect{\star}}{\frameSAbb}$ , is used if a vector is not represented in its original frame.

\subsection{Human Operator}\label{sec:model_human}
Let $\humanPos\in\nR{3}$ and $\humanVel\in\mathbb{R}^3$, both expressed in $\frameM$, denote the position and velocity of $\originH$ with respect to $\originM$, i.e. the haptic device end-effector position and velocity with respect to the idle configuration.
Similarly, we define the attitude and angular rate of $\frameH$ with respect to $\frameM$ as $\humanAtt\in\SO{3}$ and $\humanRate\in\mathbb{R}^3$, the latter expressed in $\frameH$. 
Since the human is in constant contact with the end-effector, its pose and twist values are similar to the ones of the end-effector.
The dynamic relation between these values is modeled in $\frameH$ as:
\begin{align}
    \humanInert\matrice{\vecInFrame{\dhumanVel}{\frameHAbb}\\\dhumanRate}
    + \humanDamping\matrice{\vecInFrame{\humanVel}{\frameHAbb}\\\humanRate}
    = -\humanWrench + \humanWrenchActive,
    \label{eq:human_model}
\end{align}
where $\humanWrenchActive\in\nR{6}$ are wrenches \add{(stacked forces and torques)} from the muscles and $\humanWrench\in\nR{6}$ are the interaction wrenches with the haptic device.
The human inherent inertia and damping are denoted as $\humanInert\in\nR{6\times6}$ and $\humanDamping\in\nR{6\times6}$.\remove{, respectively.}

Following standard practice in both \ac{RC} and manned rotary-wing vehicle piloting, it is assumed that the visual frame of the human is identical to $\frameS$.
In fact, the robot is commonly equipped with an onboard camera that streams images back to the operator during remote operations.
However, the human's point of view can be changed as desired by modifying the coordinate frame conventions used in Section~\ref{sec:teleop} accordingly.

\subsection{Haptic Device}\label{sec:model_master}
Since the inertia and damping of the human dynamics in \eqref{eq:human_model} are generally unknown, the haptic device manipulator is interacting with an unknown environment with low impedance.
As explained in detail in \cite{2020-Zimmermann}, this is generally undesired for torque-controlled systems, since the contact constraints can not be accurately described.
To still ensure compliant interaction and haptic transparency, an admittance filter is introduced and combined with a low-level joint position controller.
Assuming perfect tracking, the closed-loop robot arm dynamics can then be approximated as:
\begin{align}
    \humanAdmInertia\matrice{\vecInFrame{\dhumanVel}{\frameHAbb}\\\dhumanRate}
    + \humanAdmDamp\matrice{\vecInFrame{\humanVel}{\frameHAbb}\\\humanRate}
    =
    \humanWrench + \feedbackTotalWrench,
    \label{eq:panda_model}
\end{align}
with inertia $\humanAdmInertia=diag(\humanAdmInertiaTransl,\humanAdmInertiaRot)\in\nR{6\times6}$ and damping $\humanAdmDamp=diag(\humanAdmDampTransl,\humanAdmDampRot)\in\nR{6\times6}$ tuning matrices and where $\feedbackTotalWrench\in\nR{6}$ are the desired feedback wrenches to be applied to the user (see Section~\ref{ssec:teleop_fb_gen}).

\subsection{OMAV}\label{sec:model_slave}
We denote the position and velocity of the \ac{OMAV}'s center-of-mass $\originS$ with respect to $\frameW$ with $\slavePos\in\mathbb{R}^3$ and $\slaveVel\in\mathbb{R}^3$.
The orientation and angular rate of $\frameS$ with respect to $\frameW$ is given as $\slaveAtt\in \SO{3}$ and $\slaveRate\in\mathbb{R}^3$, the latter expressed in $\frameS$.
To allow for compliant interaction, we assume that the robot is controlled by an impedance controller similar to the one presented in \cite{2020-Bodie}.
Thus, the rendered closed-loop dynamics in $\frameS$ are
\begin{align}
    \slaveVirtInert\matrice{\vecInFrame{\slaveAcc}{\frameSAbb}\\\slaveAngAcc}
    + \slaveVirtDamp\matrice{\slaveErrorVel\\ \slaveErrorRate}
    + \slaveVirtStiff\matrice{\slaveErrorPos\\ \slaveErrorAtt}
    = \slaveExtWrench.
    \label{eq:omav_impedance}
\end{align}
The virtual inertia $\slaveVirtInert\in\nR{6\times6}$, damping $\slaveVirtDamp\in\nR{6\times6}$ and stiffness $\slaveVirtStiff\in\nR{6\times6}$ are tuning parameters of the on-board controller and $\slaveExtWrench\in\nR{6}$ describes external disturbances acting on the platform.
In the context of this work, we assume that such external disturbances originate solely from physical interaction of the robot with the environment.

Given a desired position $\slaveRefPos\in\mathbb{R}^3$ and velocity $\slaveRefVel\in\mathbb{R}^3$ in $\frameW$, attitude $\slaveRefAtt\in \SO{3}$ and angular rate $\slaveRefRate\in\mathbb{R}^3$ in $\frameRefS$ of the \ac{OMAV}, the tracking errors are defined in $\frameS$ as
\begin{align}
    \slaveErrorPos &= \slaveAtt\transp\left({\slavePos} - {\slaveRefPos}\right)\\
    \slaveErrorAtt &= \frac{1}{2}(\slaveRefAtt\transp\slaveAtt - \slaveAtt\transp\slaveRefAtt)^\vee\\
    \slaveErrorVel &= \slaveAtt\transp\left(\slaveVel - \slaveRefVel\right)\\
    \slaveErrorRate &= \slaveRate - \slaveAtt\transp\slaveRefAtt\slaveRefRate,
\end{align}
where the \textit{vee}-map $(\cdot)^\vee:\mathfrak{so}{(3)}\rightarrow\nR{3}$ is the inverse of the skew-symmetric operator $[\cdot]_\times:\nR{3}\rightarrow\mathfrak{so}{(3)}$.

\section{TELEOPERATION}\label{sec:teleop}
An overview of the omnidirectional bilateral teleoperation framework proposed in this work is presented in \fig\ref{fig:system_design_detail}.
The employed reference and feedback generation policies are explained in detail in the following sections.
\subsection{Rate Control Reference Generation}\label{ssec:teleop_ref_gen}
Rate control is a well established method to teleoperate aerial vehicles, since it provides an intuitive mapping between the restricted input workspace and the potentially infinite robot workspace \cite{2005-ConKha}. 
However, in the state-of-the-art literature, it is often limited to translation only due to the underactuated nature of standard fixed-rotor \acp{MAV}.
This restriction does not apply to \acp{OMAV}, which is why the concept is extended in this work to include rotational rate control as well.
Essentially, any translational or rotational deviation between $\frameH$ and $\frameM$ is translated into a corresponding translational velocity or angular rate reference for the robot.
Based on this, the translational references are computed as
\begin{subequations}
\begin{align}
    \vecInFrame{\slaveRefVel}{\frameSAbb} &= \refgenMaxVel\humanPos\\
    \vecInFrame{\slaveRefPos}{\frameSAbb} &= \int_0^t \vecInFrame{\slaveRefVel}{\frameSAbb}(s) ds,
\end{align}
    \label{eq:ref_gen_transl}
\end{subequations}
and similarly for rotation
\begin{subequations}
\begin{align}
    \vecInFrame{\slaveRefRate}{\frameSAbb} &= \frac{\refgenMaxRate}{2}\left(\humanAtt-\humanAtt^\top\right)^\vee\\
    \slaveRefAtt &= \int_0^t \slaveRefAtt(s)[\slaveRefRate(s)]_{\times} ds,
\end{align}
    \label{eq:ref_gen_rot}
\end{subequations}
where $\refgenMaxVel$ and $\refgenMaxRate$ are used to tune how fast the vehicle moves.
In accordance with the assumption of the human's point of view from an onboard camera, all references are provided in the body frame $\frameS$.

\subsection{Feedback Generation}\label{ssec:teleop_fb_gen}
As stated in Section~\ref{sec:intro}, the design of adequate feedback wrenches is crucial to ensure ease of operation and situational awareness for the human operator.
Hereby, the overall feedback wrench $\feedbackTotalWrench\in\nR{6}$ expressed in $\frameM$ is often a combination of multiple contributions, each one representing a different aspect of the current task (e.g. object avoidance, guiding towards a waypoint).
In the context of this work, we restrict our analysis to the well established recentering $\feedbackRecenterWrench\in\nR{6}$ and interaction wrench $\feedbackExtWrench\in\nR{6}$ feedback.
Eventually, the total feedback wrench takes the form
\begin{align}
    \feedbackTotalWrench=
    \feedbackRecenterWrench+
    \feedbackExtWrench,
    \label{eq:fb_gen_tot}
\end{align}
with $\feedbackRecenterWrench$ and $\feedbackExtWrench$ computed as explained below.

\subsubsection{Recentering Wrench \texorpdfstring{$\feedbackRecenterWrench$}{TEXT}} \label{sec:Recentering Wrench}
When using rate control reference generation, the most essential type of feedback is the \textit{recentering}.
The recentering wrench $\feedbackRecenterWrench$ in $\frameM$ aims to move the haptic device's end-effector back to its idle pose, in other words make $\frameH$ identical to $\frameM$:
\begin{align}
    \feedbackRecenterWrench =
    - \feedbackRecenterStiff\matrice{\humanPos\\\frac{1}{2}\left(\humanAtt-{\humanAtt}^\top\right)^\vee},
    \label{eq:fg_gen_recenter}
\end{align}
where the stiffness $\feedbackRecenterStiff=diag(\feedbackRecenterStiffTransl,\feedbackRecenterStiffRot)\in\nR{6\times6}$ is a tuning parameter.
Under rate control, recentering the end-effector will cause the robot to slow down and eventually hold position and attitude.
In that sense, adding a virtual spring on the human side translates to the addition of a virtual damper on the robot side.
Without recentering feedback, it would be almost impossible to manually zero the haptic device in all six directions and achieve static hover.
Additionally, it allows the human operator to let go of the handle at any time and the robot will automatically stabilize at its current pose.

When targeting applications involving physical contact with the environment however, the recentering wrench is no longer sufficient to ensure situational awareness.
Especially in cases where the camera view might not be conclusive about whether the robot is in contact or not, additional interaction-specific information must be provided.

\subsubsection{Interaction Wrench \texorpdfstring{$\feedbackExtWrench$}{TEXT}}
In state-of-the-art literature, interaction specific feedback involves reflecting the measured or estimated forces at the contact point $\contactForces\in\nR{3}$ back to the operator, i.e. the forces being applied by the environment to the aerial robot expressed in $\frameS$. 
This work proposes an extension for omnidirectional vehicles, whereby the torques $\contactTorques\in\nR{3}$ in $\frameS$ acting on the vehicle at the contact point are also included.
Considering the offset between the vehicle's center-of-mass and the part of it that is in contact (i.e. the \textit{tool}) $\bm{r}_{\originS T}\in\nR{3}$ in $\frameS$, the interaction wrench feedback is then given as:
\begin{align}
    \feedbackExtWrench =
    \matrice{\contactForces\\\contactTorques} +
    \matrice{\vZero\\\bm{r}_{\originS T}\times\contactForces},
    \label{eq:interaction_wrench}
\end{align}
where $\feedbackExtWrench$ is expressed in $\frameM$, again using the assumption that $\frameM$ is aligned with $\frameS$.
Notice that this corresponds to the external wrench $\slaveExtWrench$ acting on the vehicle's center-of-mass during interaction, effectively representing the same wrench a human being would feel when holding the tool.

\subsection{Stability Considerations}
In this paper, no formal proof of the stability of the teleoperation system is provided.
While this will be the focus of future work, some stability-related aspects are briefly discussed here.
The structure of our proposed framework, namely rate control in combination with recentering and environment force feedback, has some similarity with the scheme presented in \cite{2011-Rifai}.
In that paper, stability of the teleoperation loop was proven when subject to bounded operator and external forces.
Although an underactuated system is considered, that proof suggests the existence of similar formal guarantees for the case of omnidirectional vehicles.
Furthermore, the flight experiments presented in the following section already verify the practical stability of the developed teleoperation policy.

\section{RESULTS}\label{sec:results}
\begin{figure*}[hbt]
    \begin{center}
        \includegraphics[width=\linewidth]{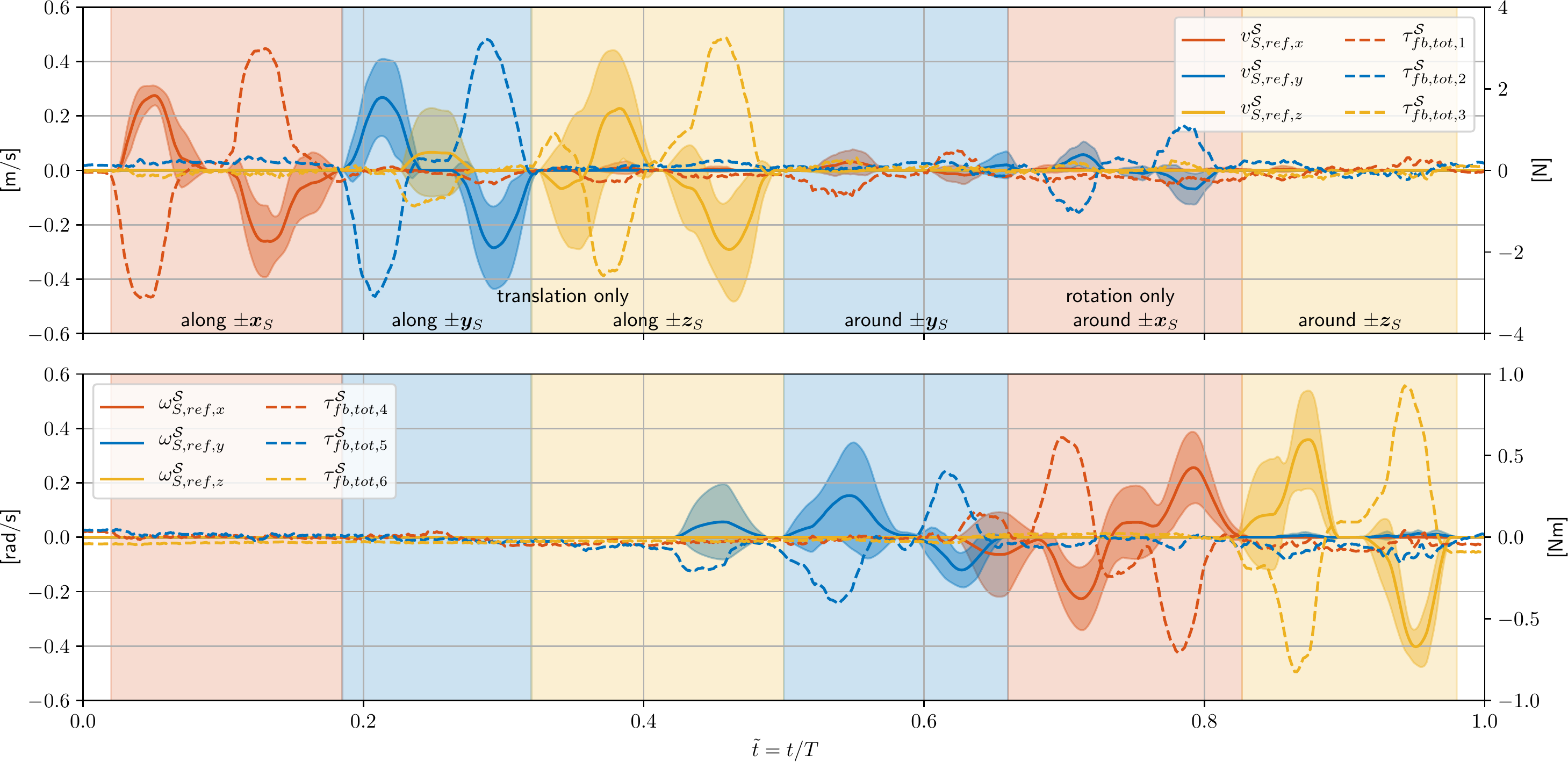}
    \end{center}
    \caption{\add{Decoupled translational (top) and rotational (bottom) reference generation. Solid lines and corresponding shaded areas indicate the mean and standard deviation of the operator inputs, respectively and dashed lines indicate the mean total feedback wrench over five repeated trials. Coupling effects related to unintended input agitation despite recentering wrench can be observed, especially $\yS, \zS$ coupling during translation inputs and $\xS, \yS$ translation during rotation inputs.}}
    \label{fig:eval_omnidir}
\end{figure*}
\subsection{Experimental Setup}
Indoor flight experiments are conducted to evaluate the capabilities and performance of the proposed bilateral teleoperation setup.
The employed system is shown in Fig.~\ref{fig:frames}, consisting of a 7DoF Franka Emika Panda arm with a handle attached to the end-effector for a haptic device and the \ac{OMAV} based on the tiltrotor aerial platform introduced in \cite{2020-Allenspach}.
A measure of the interaction force between the human and the robot arm is obtained at $\SI{800}{\hertz}$ with a 6-axis Rokubi force-torque sensor mounted between the end-effector and the handle.
The Panda arm is running the default cartesian velocity controller in combination with an admittance filter, effectively rendering the closed-loop dynamics in \eqref{eq:panda_model}.
\add{The admittance gains are set as low as possible to improve haptic transparency, while still ensuring a minimum dissipation to maintain system stability. Reference generation and recentering parameters are tuned to render robot velocities and feedback wrenches comfortable for the user.
The specific values used during experiments are listed in Table~\ref{tab:exp_values}.}
It should be noted that the framework can easily be combined with other types of haptic devices, given they support 6DoF motion input and wrench feedback rendering.

\begin{table}[t]
    \centering
    \add{
    \begin{tabular}{ll|ll}
       Parameter & & Value &\\\midrule
       $\humanAdmInertiaTransl$ & $\humanAdmInertiaRot$ & $10\eye{3\times3}[\si{\kilo\gram}]$ & $\eye{3\times3}[\si{\kilo\gram\metre^2}]$\\
       $\humanAdmDampTransl$ & $\humanAdmDampRot$ & $5\eye{3\times3}[\si{\kilo\gram\per\second}]$ & $\eye{3\times3}[\si{\kilo\gram\second^{-1}\metre^2}]$\\
       $\refgenMaxVel$ & $\refgenMaxRate$ &
           $1[\si{\per\second}]$ & $1[\si{\per\second}]$
       \\
       $\feedbackRecenterStiffTransl$ & $\feedbackRecenterStiffRot$ &
           $50\eye{3\times3}[\si{\newton\per\metre}]$ & $2\eye{3\times3}[\si{\newton\metre}]$\\
    \end{tabular}
    }
    \caption{\add{System parameters for experiments.}}
    \label{tab:exp_values}
\end{table}

The remote environment is equipped with a \ac{MOCAP}, providing pose measurements for the \ac{OMAV} at $\SI{100}{\hertz}$.
An EKF-based state estimator provides the full state estimate $\slavePos,\slaveVel,\slaveAtt,\slaveRate$ required by the impedance controller, fusing \ac{MOCAP} with onboard IMU data (accelerometer and gyroscope). \add{The controller is tuned according to \cite{2020-Bodie}.}
\add{A safety tether is connected to the \ac{OMAV} but kept slack to limit external disturbances.}
\remove{As visible in Fig., }Additionally, a rigid rod of approximately $\SI{0.6}{\metre}$ length with a soft ball at the end is attached to the robot and acts as the interaction tool for contact-based tasks.
The rod provides sufficient clearance for the propellers to allow successful interaction without the risk of collision.
The ball acts as a mechanical damper to soften hard impacts.
Interaction forces are measured by a 6-axis Rokubi force-torque sensor mounted between the \ac{OMAV} and the rod.
Hereby, the sensor data is transformed accordingly to represent forces and torques acting on the vehicle center-of-mass (see Section~\ref{ssec:teleop_fb_gen}).

Note that the human is directly looking at the robot, instead of viewing from the robot perspective as mentioned in earlier sections.
However, their visual frame is still well aligned with $\frameS$, since the haptic device is placed directly behind the robot and the experiments only involve small attitude changes ($<\SI{10}{\degree}$).

\subsection{Translational and Rotational Reference Generation}

Recall that the aim of this paper is to analyze the suitability of standard bilateral teleoperation methodologies when extended to 6DoF for omnidirectional vehicles.
As a first criteria, the operator must be able to generate decoupled motion commands in all translational and rotational DoF, in order to fully exploit the omnidirectional capabilities of the \ac{OMAV}.
This requires the human 
to accurately render decoupled forces and torques at a single interaction point, namely the handle of the haptic device.
Thus, evaluating whether they are physically and cognitively capable of performing such manipulation is crucial for the feasibility of the proposed framework.
This is tested experimentally by \add{repeatedly} tasking an operator with sequentially actuating each individual input axis without introducing motion in other directions.
\add{The resulting translational and rotational reference velocity statistics from five trials with a single operator over the experiment duration $T$, as well as the total feedback wrench acting on the handle are shown in Fig.~\ref{fig:eval_omnidir}.}
Note that this test only involves free flight operation, meaning that the displayed feedback wrench only consists of recentering actions, i.e. $\feedbackTotalWrench=\feedbackRecenterWrench$.
The effect of this recentering term is clearly visible, shown by the constant wrench opposing the twist commands, aiming to restore the handle's idle pose in the local environment.

The results clearly show an unintended coupling between the different axes on the input device.
During the translational reference generation along $\yS$ ($\tilde{t}\in[0.15,0.3]$) for example, non-zero velocity references in $\zS$ can be observed.
Similarly, when trying to move along $\zS$ only ($\tilde{t}\in[0.3,0.5]$), additional rotation along $\yS$ is accidentally introduced.
A similar phenomenon is observed when the user is tasked with performing rotations only.
These coupling effects \remove{strongly }worsen the performance when precise maneuvering is required, such as for high-accuracy tasks or when operating in confined spaces.

In summary, it appears that producing decoupled reference inputs, especially a pure rotation at the handle, are physically challenging for the operator, despite the supporting recentering wrench.
Making the recentering gain adaptive could help to constrain the user input to a single axis at a time, by making the remaining axes more stiff.
Hereby, \add{the adaptive solution must still allow} full exploitation of the omnidirectional capabilities.
A detailed study of such methodologies is left for future work.

\subsection{Push-and-Slide Interaction}
\begin{figure}
    \begin{center}
        \includegraphics[trim={0cm 7cm 0cm 3cm},clip,width=\linewidth]{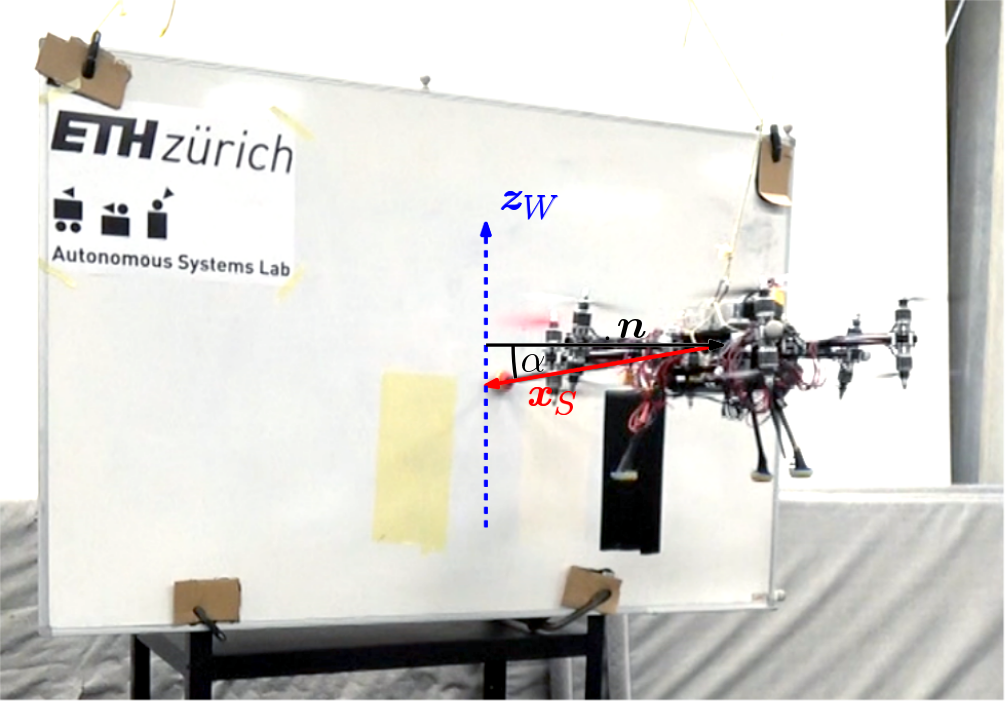}
    \end{center}
    \caption{Experimental setup for vertical sliding experiment along $\zW$. Uncompensated friction opposing the motion results in an angular offset $\alpha$ between the surface norm $\bm{n}$ and the body x-axis $\xS$.} 
    \label{fig:exp_sliding}
\end{figure}

\begin{figure}
    \begin{center}
        \includegraphics[width=\linewidth]{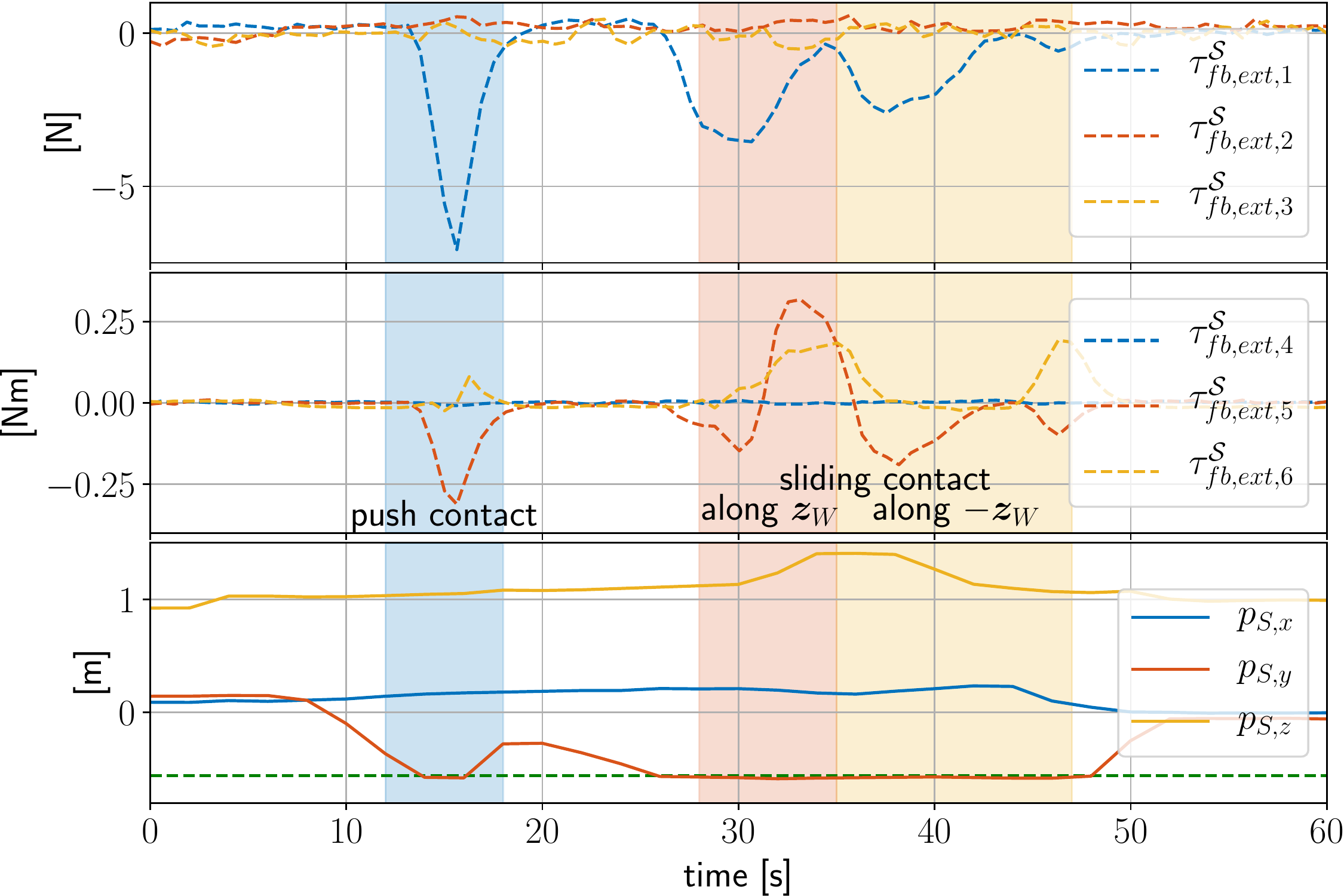}
    \end{center}
    \caption{Push-and-slide contact experiment. Contact with the wall (\remove{at $\slavePosY=\SI{-0.58}{\metre}$, }the dashed line) and sliding along $\pm\zW$ are shown. External torques due to friction effects (see Fig.~\ref{fig:exp_sliding}) can be observed.}
    \label{fig:eval_interaction}
\end{figure}

Apart from omnidirectional reference generation in contact-free conditions, a bilateral teleoperation framework for an \ac{OMAV} must allow the operator to perform physical interaction tasks as well.
We evaluate this requirement by performing a push-and-slide operation with a whiteboard, as shown in Fig.~\ref{fig:exp_sliding}.
This is a common task in contact-based inspection applications.
During the first phase of the experiment, the operator is asked to approach in a direction normal to the whiteboard surface and \remove{merely }push against it.
In a second phase, once contact with the board is established, the user is tasked with sliding along $\zW$.
The resulting interaction wrench being fed back to the user is shown in Fig.~\ref{fig:eval_interaction}.
Notice that the recentering wrench is not included here, since it is not the primary focus of the experiment.
Additionally, the position of the \ac{OMAV} in $\frameW$ is visualized to highlight the motion with respect to the whiteboard surface.

When pushing against the board (highlighted in blue), a clear spike in $-\xS$ feedback force can be observed, indicating the presence and intensity of the contact to the operator.
The non-zero torque around $-\yS$ originates from the second term in \eqref{eq:interaction_wrench} and is caused by a misalignment between the surface normal $\bm{n}$ and $\xS$ (see Fig.~\ref{fig:exp_sliding} with $\alpha<0$), resulting in an external pitching torque.
During vertical sliding, friction effects acting at the tool tip cause the same behavior.
While moving upwards (highlighted in orange), the tool lags behind due to the high friction force. Since the tool and the connecting rod are rigidly attached to the \ac{OMAV}, this causes the vehicle to pitch down slightly, producing the observed positive feedback torque around $\yS$.
The opposite behavior occurs when sliding downwards (highlighted in yellow).

Compared to the omnidirectional reference generation, no immediate limitation was detected when using the proposed extension of the standard interaction force feedback.
That being said, different experiments showed a degradation in the performance of the flight controller, such as oscillations or tool skipping, in the presence of large magnitude interaction and friction forces (see also complementary video).  
In this regard, the effectiveness of the provided feedback could be improved further by including information about the robot state and its limitations.


\section{CONCLUSION} \label{sec:conclusion}
This work investigates the suitability of standard bilateral teleoperation methods in the context of \acp{OMAV}.
Based on a straightforward extension of the well established rate control, recentering and interaction force feedback policy, a bilateral teleoperation framework for an omnidirectional aerial robot has been designed and evaluated.
The human is in contact with the handle of a haptic device and performs small-scale deviations from the idle pose, thereby generating twist commands for the robot.
Wrench feedback is provided to the operator, on the one hand recentering the handle to restore its idle pose and on the other hand reflecting external forces acting on the vehicle when in contact with the environment.

Practical experiments including contact-free flight, as well as push-and-slide operation during physical interaction are conducted to evaluate the potential of the proposed approach.
Although the operator is able to control all six axes of the \ac{OMAV}, performing decoupled motion in a single DoF only is practically challenging.
This is a fundamental issue of the straightforward extension of standard methodologies on this new types of platforms, which shows the need for additional measures to suppress unintended inputs.
Being able to prevent coupling effects is absolutely mandatory, especially when precise maneuvering is required for high-accuracy tasks and in confined spaces.
Future work will focus on addressing this problem through the use of adaptive axes stiffness and human intention detection.

In summary, effective bilateral teleoperation for \acp{OMAV} cannot be achieved by simple extension of standard teleoperation methodologies but rather requires more sophisticated policies to fully exploit the capabilities of these vehicles.



\addtolength{\textheight}{-10cm}
\bibliographystyle{IEEEtran}
\bibliography{./bibAlias,./bibCustom}

\begin{thebibliography}{10}
\providecommand{\url}[1]{#1}
\csname url@samestyle\endcsname
\providecommand{\newblock}{\relax}
\providecommand{\bibinfo}[2]{#2}
\providecommand{\BIBentrySTDinterwordspacing}{\spaceskip=0pt\relax}
\providecommand{\BIBentryALTinterwordstretchfactor}{4}
\providecommand{\BIBentryALTinterwordspacing}{\spaceskip=\fontdimen2\font plus
\BIBentryALTinterwordstretchfactor\fontdimen3\font minus
  \fontdimen4\font\relax}
\providecommand{\BIBforeignlanguage}[2]{{%
\expandafter\ifx\csname l@#1\endcsname\relax
\typeout{** WARNING: IEEEtran.bst: No hyphenation pattern has been}%
\typeout{** loaded for the language `#1'. Using the pattern for}%
\typeout{** the default language instead.}%
\else
\language=\csname l@#1\endcsname
\fi
#2}}
\providecommand{\BIBdecl}{\relax}
\BIBdecl

\bibitem{2019e-TogTelGasSabBicMalLanSanRevCorFra}
M.~Tognon, H.~A. {Tello~Ch\'avez}, E.~Gasparin, Q.~Sabl\'e, D.~Bicego,
  A.~Mallet, M.~Lany, G.~Santi, B.~Revaz, J.~Cort\'es, and A.~Franchi, ``A
  truly redundant aerial manipulator system with application to push-and-slide
  inspection in industrial plants,'' \emph{IEEE Robotics and Automation
  Letters}, vol.~4, no.~2, pp. 1846--1851, 2019.

\bibitem{2019-BodBruPanWalPfaAngSieNie}
K.~Bodie, M.~Brunner, M.~Pantic, S.~Walser, P.~Pf{\"a}ndler, U.~Angst,
  R.~Siegwart, and J.~Nieto, ``An omnidirectional aerial manipulation platform
  for contact-based inspection,'' in \emph{Proceedings of Robotics: Science and
  Systems}, FreiburgimBreisgau, Germany, June 2019.

\bibitem{2021-Ollero}
A.~Ollero, M.~Tognon, A.~Suarez, D.~Lee, and A.~Franchi,
  ``\href{https://ieeexplore.ieee.org/stamp/stamp.jsp?tp=&arnumber=9462539}{Past,
  Present, and Future of Aerial Robotic Manipulators},'' \emph{IEEE
  Transactions on Robotics}, vol.~38, no.~1, pp. 626--645, 2022.

\bibitem{2018-Ruggiero}
F.~{Ruggiero}, V.~{Lippiello}, and A.~{Ollero},
  ``\href{https://ieeexplore.ieee.org/document/8299552}{Aerial Manipulation: A
  Literature Review},'' \emph{IEEE Robotics and Automation Letters}, vol.~3,
  no.~3, pp. 1957--1964, 2018.

\bibitem{2020-Meng}
X.~Meng, Y.~He, and J.~Han,
  ``\href{https://www.cambridge.org/core/journals/robotica/article/survey-on-aerial-manipulator-system-modeling-and-control/5FDD2404D65EF73477CF10CA62B69720}{Survey
  on Aerial Manipulator: System, Modeling, and Control},'' \emph{Robotica},
  vol.~38, no.~7, p. 1288–1317, 2020.

\bibitem{2021-BodTogSie}
K.~Bodie, M.~Tognon, and R.~Siegwart, ``Dynamic end effector tracking with an
  omnidirectional parallel aerial manipulator,'' \emph{IEEE Robotics and
  Automation Letters}, vol.~6, no.~4, pp. 8165--8172, 2021.

\bibitem{2018-ParLeeAhnKimHerYanLee}
S.~{Park}, J.~{Lee}, J.~{Ahn}, M.~{Kim}, J.~{Her}, G.~{Yang}, and D.~{Lee},
  ``Odar: Aerial manipulation platform enabling omnidirectional wrench
  generation,'' \emph{IEEE/ASME Transactions on Mechatronics}, vol.~23, no.~4,
  pp. 1907--1918, 2018.

\bibitem{2018-BreDAn}
D.~Brescianini and R.~{D'Andrea}, ``An omni-directional multirotor vehicle,''
  \emph{Mechatronics}, vol.~55, pp. 76--93, 2018.

\bibitem{2019-Ryll}
M.~Ryll, G.~Muscio, F.~Pierri, E.~Cataldi, G.~Antonelli, F.~Caccavale,
  D.~Bicego, and A.~Franchi,
  ``\href{https://journals.sagepub.com/doi/full/10.1177/0278364919856694#articleCitationDownloadContainer}{6D
  interaction control with aerial robots: The flying end-effector paradigm},''
  \emph{The International Journal of Robotics Research}, vol.~38, no.~9, pp.
  1045--1062, 2019.

\bibitem{2019-Trujillo}
M.~A. Trujillo, J.~R. Martínez-de Dios, C.~Martín, A.~Viguria, and A.~Ollero,
  ``\href{https://www.mdpi.com/1424-8220/19/6/1305}{Novel Aerial Manipulator
  for Accurate and Robust Industrial NDT Contact Inspection: A New Tool for the
  Oil and Gas Inspection Industry},'' \emph{Sensors}, vol.~19, no.~6, 2019.

\bibitem{2020-Bodie}
K.~Bodie, M.~Brunner, M.~Pantic, S.~Walser, P.~Pfandler, U.~Angst, R.~Siegwart,
  and J.~Nieto, ``\href{https://arxiv.org/pdf/2003.09516.pdf}{Active
  Interaction Force Control for Contact-Based Inspection With a Fully Actuated
  Aerial Vehicle},'' \emph{IEEE Transactions on Robotics}, pp. 1--14, 2020.

\bibitem{2014-MerStrCar}
A.~Y. {Mersha}, S.~{Stramigioli}, and R.~{Carloni}, ``On bilateral
  teleoperation of aerial robots,'' \emph{IEEE Transactions on Robotics},
  vol.~30, no.~1, pp. 258--274, 2014.

\bibitem{2010-Schill}
F.~Schill, X.~Hou, and R.~Mahony,
  ``\href{https://www.researchgate.net/publication/265518881_Admittance_mode_framework_for_haptic_teleoperation_of_hovering_vehicles_with_unlimited_workspace}{Admittance
  mode framework for haptic teleoperation of hovering vehicles with unlimited
  workspace},'' in \emph{2010 Australasian Conf. on Robotics \& Automation,
  (Brisbane, Australia)}, 12 2010.

\bibitem{2011-Rifai}
H.~Rifa\"{i}, M.-D. Hua, T.~Hamel, and P.~Morin,
  ``\href{https://reader.elsevier.com/reader/sd/pii/S1474667016458398}{Haptic-based
  bilateral teleoperation of underactuated Unmanned Aerial Vehicles},''
  \emph{IFAC Proceedings Volumes}, vol.~44, no.~1, pp. 13\,782--13\,788, 2011,
  18th IFAC World Congress.

\bibitem{2018-Masone}
C.~Masone, M.~Mohammadi, P.~R. Giordano, and A.~Franchi,
  ``\href{https://journals.sagepub.com/doi/full/10.1177/0278364918802006#articleCitationDownloadContainer}{Shared
  planning and control for mobile robots with integral haptic feedback},''
  \emph{The International Journal of Robotics Research}, vol.~37, no.~11, pp.
  1395--1420, 2018.

\bibitem{2015-Gioioso}
G.~{Gioioso}, M.~{Mohammadi}, A.~{Franchi}, and D.~{Prattichizzo},
  ``\href{https://ieeexplore.ieee.org/stamp/stamp.jsp?tp=&arnumber=7139018}{A
  force-based bilateral teleoperation framework for aerial robots in contact
  with the environment},'' in \emph{2015 IEEE International Conference on
  Robotics and Automation (ICRA)}, 2015, pp. 318--324.

\bibitem{2019-Islam}
S.~{Islam}, R.~{Ashour}, and A.~{Sunda-Meya},
  ``\href{https://ieeexplore.ieee.org/stamp/stamp.jsp?arnumber=8598945}{Haptic
  and Virtual Reality Based Shared Control for MAV},'' \emph{IEEE Transactions
  on Aerospace and Electronic Systems}, vol.~55, no.~5, pp. 2337--2346, 2019.

\bibitem{2020-Lee}
J.~Lee, R.~Balachandran, Y.~S. Sarkisov, M.~D. Stefano, A.~Coelho, K.~Shinde,
  M.~J. Kim, R.~Triebel, and K.~Kondak,
  ``\href{https://arxiv.org/abs/2003.11509}{Visual-Inertial Telepresence for
  Aerial Manipulation},'' 2020.

\bibitem{2020-Coelho}
A.~{Coelho}, H.~{Singh}, K.~{Kondak}, and C.~{Ott}, ``Whole-body bilateral
  teleoperation of a redundant aerial manipulator,'' in \emph{2020 IEEE
  International Conference on Robotics and Automation (ICRA)}, 2020, pp.
  9150--9156.

\bibitem{2020-Zimmermann}
Y.~{Zimmermann}, E.~B. {Küçüktabak}, F.~{Farshidian}, R.~{Riener}, and
  M.~{Hutter}, ``\href{https://ieeexplore.ieee.org/document/9341054}{Towards
  Dynamic Transparency: Robust Interaction Force Tracking Using Multi-Sensory
  Control on an Arm Exoskeleton},'' in \emph{2020 IEEE/RSJ International
  Conference on Intelligent Robots and Systems (IROS)}, 2020, pp. 7417--7424.

\bibitem{2005-ConKha}
F.~Conti and O.~Khatib, ``Spanning large workspaces using small haptic
  devices,'' in \emph{First Joint Eurohaptics Conference and Symposium on
  Haptic Interfaces for Virtual Environment and Teleoperator Systems. World
  Haptics Conference}, 2005, pp. 183--188.

\bibitem{2020-Allenspach}
M.~Allenspach, K.~Bodie, M.~Brunner, L.~Rinsoz, Z.~Taylor, M.~Kamel,
  R.~Siegwart, and J.~Nieto,
  ``\href{https://journals.sagepub.com/doi/full/10.1177/0278364920943654#articleCitationDownloadContainer}{Design
  and optimal control of a tiltrotor micro-aerial vehicle for efficient
  omnidirectional flight},'' \emph{The International Journal of Robotics
  Research}, vol.~39, no. 10-11, pp. 1305--1325, 2020.

\end{thebibliography}

\end{document}